\documentclass{article}
\usepackage{ijcai19}

\usepackage{times}
\usepackage{soul}
\usepackage{url}
\usepackage[hidelinks]{hyperref}
\usepackage[utf8]{inputenc}
\usepackage[small]{caption}

\usepackage{graphicx, subfigure}

\usepackage{amsmath}
\usepackage{amssymb}
\usepackage{booktabs}
\usepackage{algorithm}
\usepackage{algorithmic}

\title{Localizing Unseen Activities in Video via Image Query}

\author{
Zhu Zhang$^1$
\and
Zhou Zhao$^1$\footnote{Zhou Zhao is the corresponding author.}\and
Zhijie Lin$^1$\and
Jingkuan Song$^2$\And
Deng Cai$^3$
\affiliations
$^1$College of Computer Science, Zhejiang University, China\\
$^2$University of Electronic Science and Technology of China, China\\
$^3$State Key Lab of CAD\&CG, Zhejiang University, China\\
\emails
\{zhangzhu, zhaozhou, linzhijie\}@zju.edu.cn,
\{jingkuan.song, dengcai\}@gmail.com
}

\begin{document}

\maketitle

\begin{abstract}
Action localization in untrimmed videos is an important topic in the field of video understanding. However, existing action localization methods are restricted to a pre-defined set of actions and cannot localize unseen activities. Thus, we consider a new task to localize unseen activities in videos via image queries, named Image-Based Activity Localization. This task faces three inherent challenges: (1) how to eliminate the influence of semantically inessential contents in image queries; (2)  how to deal with the fuzzy localization of inaccurate image queries; (3) how to determine the precise boundaries of target segments. We then propose a novel self-attention interaction localizer to retrieve unseen activities in an end-to-end fashion. Specifically, we first devise a region self-attention method with relative position encoding to learn fine-grained image region representations. Then, we employ a local transformer encoder to build multi-step fusion and reasoning of image and video contents. We next adopt an order-sensitive localizer to directly retrieve the target segment. Furthermore, we construct a new dataset ActivityIBAL by reorganizing the ActivityNet dataset. The extensive experiments show the effectiveness of our method.
\end{abstract}

\section{Introduction}
\begin{figure}[t]
\centering
\includegraphics[width=0.47\textwidth]{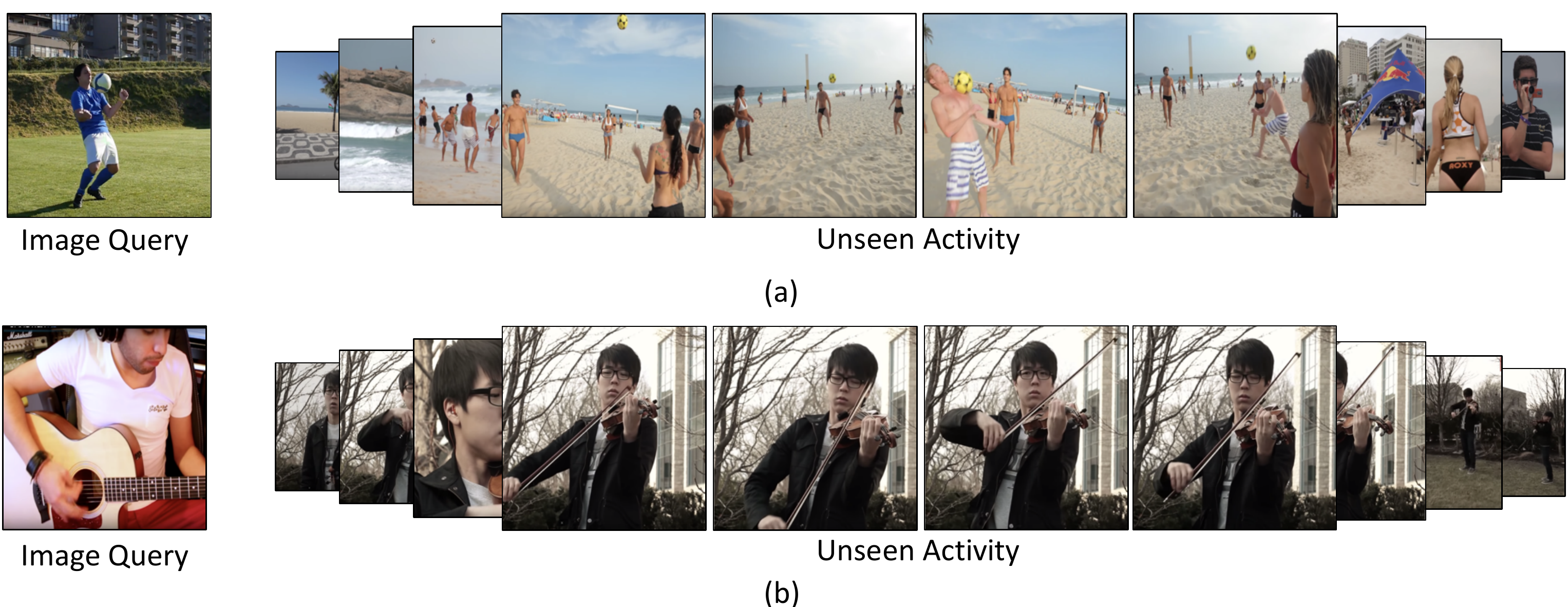}
\caption{Image-Based Activity Localization.}\label{fig:exp}
\end{figure}

With the rapid growth of video data, action understanding has attracted much attention in the computer vision community.
Early works mainly focus on action recognition~\cite{simonyan2014two,tran2015learning} of trimmed video clips that only contain one activity. 
Recently, some works~\cite{yeung2016end,shou2016temporal,chao2018rethinking} have made great progress in action localization, which retrieves multiple activities from a long, untrimmed video and meanwhile predicts the temporal boundaries of these activities.
However, existing action localization methods are restricted to a pre-defined set of actions.
Although the action set can be relatively large~\cite{caba2015activitynet}, it is still difficult to cover various and complex activities in the real world. 
Thus, in practical applications, these approaches are hard to recognize those unseen activities, which do not appear in the training data.
But if there is an image query as a reference, we can localize the unseen activities by matching visual contents with respect to objects, appearances, motions and interactions.
As shown in Figure~\ref{fig:exp}(a), the activity ``people play soccer on the beach'' has the similar semantic contents with the image query, including the object ``soccer'' and the interaction between ``people'' and ``soccer''.

Motivated by these observations, we consider a new task to localize unseen activities in video via image queries, named Image-Based Activity Localization. Specifically, the inputs of this task are one untrimmed video and one image query. The image query depicts an activity that users are interested in. And the video contains one segment semantically corresponding to this activity. Then, we aim to identify the start and end boundaries of the targeted segment, even if this activity has never appeared in training data before.

Localizing unseen activities via image queries is a very challenging task, which faces three inherent difficulties. One is how to eliminate the influence of semantically inessential contents in image queries, such as playing soccer ``on the grass'' or ``on the beach'' in Figure~\ref{fig:exp}(a). It requires the precise image representation learning and video-based attention method to highlight the crucial query information. 
Recently, the self-attention mechanism~\cite{vaswani2017attention,wang2018non} has been applied to capture semantic dependencies in sequential data.
Inspired it, we devise a novel region self-attention method with relative position encoding to learn fine-grained image region representations, where the basic region features are extracted using Faster R-CNN~\cite{ren2015faster}.
Subsequently, we can perform the frame-level attention mechanism to focus on important regions and ignore unrelated parts.

The second obstacle of this task is how to deal with the fuzzy localizing of inaccurate image queries. Sometimes users may not be able to precisely describe the activities they intend to localize, and only provide an image with close semantics. As shown in Figure~\ref{fig:exp}(b), users aim to retrieve the activity ``playing the violin'' but give an image query about ``playing the guitar''. It requires the multi-level matching of both accurate and rough semantics.
Following the transformer architecture~\cite{vaswani2017attention}, we employ a local transformer encoder with multiple stacked layers to build multi-step fusion and reasoning of image and video contents, which sufficiently explores the query semantics, regardless of the accurate ``playing the guitar'' or the rough ``playing the instrument''.
Different from the original transformer, we devise a multi-head local self-attention sublayer to focus on local relation modeling, and apply a more informative fusion sublayer for query understanding.

The last difficulty of this task is how to determine the precise boundaries of target segments, which are often indistinct in videos.
We consider leveraging the contextual information to boost the localization accuracy.
Concretely, we employ an order-sensitive localizer to directly retrieve temporal boundaries, where forward and backward context is incorporated for the prediction of start and end boundaries, respectively.

Furthermore, we construct a new dataset ActivityIBAL for image-based activity localization by reorganizing the ActivityNet~\cite{caba2015activitynet}, where we split 200 action classes into three parts: 160 classes for training, 20 classes for validation and the remaining 20 classes for testing. This setting guarantees that the videos of a class used in testing do not appear during the training process.

In summary, the main contributions of this paper are as follows:
\begin{itemize}
\item We introduce a new task, named Image-based Activity Localization, which aims to localize unseen activities in untrimmed videos via an image query.
\item We propose a self-attention interaction localizer (SAIL) to deal with this challenging task. We first devise region self-attention method to learn fine-grained image region representations, and then employ a local transformer encoder to build multi-step fusion and reasoning of image and video contents. Finally, we adopt an order-sensitive localizer to directly retrieve the target segment. 
\item We reorganize the ActivityNet dataset to construct a new dataset ActivityIBAL for image-based activity localization, and validate the effectiveness of our SAIL method through extensive experiments.
\end{itemize}

\section{Related Work}
In this section, we briefly review some related work on action localization.

Temporal action localization aims to detect the action instances in untrimmed videos. 
% Existing action localization approaches can be categorized into the supervised methods and non-fully supervised ones.
In the fully-supervised fashion, some works~\cite{yeung2016end,singh2016multi} apply recurrent neural networks to capture the temporal dynamics of video contents. 
\cite{yeung2016end} propose an RNN-based agent to observe video frames and decide both where to look next and when to emit a prediction.
\cite{singh2016multi} devise two additional streams on motion and appearance for fine-grained action detection.
With the development of 3D convolution~\cite{tran2015learning},
\cite{shou2016temporal} employ three segment-based 3D ConvNets to explicitly consider temporal overlap in videos. 
\cite{xu2017r} directly encodes the video streams using a 3D fully convolutional network without pre-defined segments.
And \cite{shou2017cdc} employ temporal upsampling and spatial downsampling operations simultaneously.
Furthermore, \cite{zhao2017temporal} model the temporal structure of each action instance via a temporal pyramid.
\cite{lin2017single} skip the proposal generation and directly detect action instances based on temporal convolutional layers.
And inspired by the Faster R-CNN object detection framework~\cite{ren2015faster}, \cite{chao2018rethinking} develop an improved method for temporal action localization. 

In the non-fully supervised setting, \cite{wang2017untrimmednets}  directly learn action recognition from untrimmed videos without the temporal action annotations. 
And \cite{nguyen2018weakly} identify a sparse subset of key segments associated with target actions and fuse them through adaptive temporal pooling.
Moreover, \cite{soomro2017unsupervised} learn unsupervised action localization as a Knapsack problem.

% Although these action localization works have achieved promising performance, they still are restricted to a pre-defined set of actions and cannot localize unseen activities. 
To go beyond the pre-defined set of actions, \cite{gao2017tall,hendricks2017localizing,chen2019localizing,zhang2019crossmodal} focus on temporal localization of actions by natural language queries.
These methods can localize complex actions according to sentence queries, but are still difficult to recognize unseen activities.
Thus, we consider localizing unseen activities in untrimmed videos via an image query.

\begin{figure*}[t]
\centering
\includegraphics[width=1\textwidth]{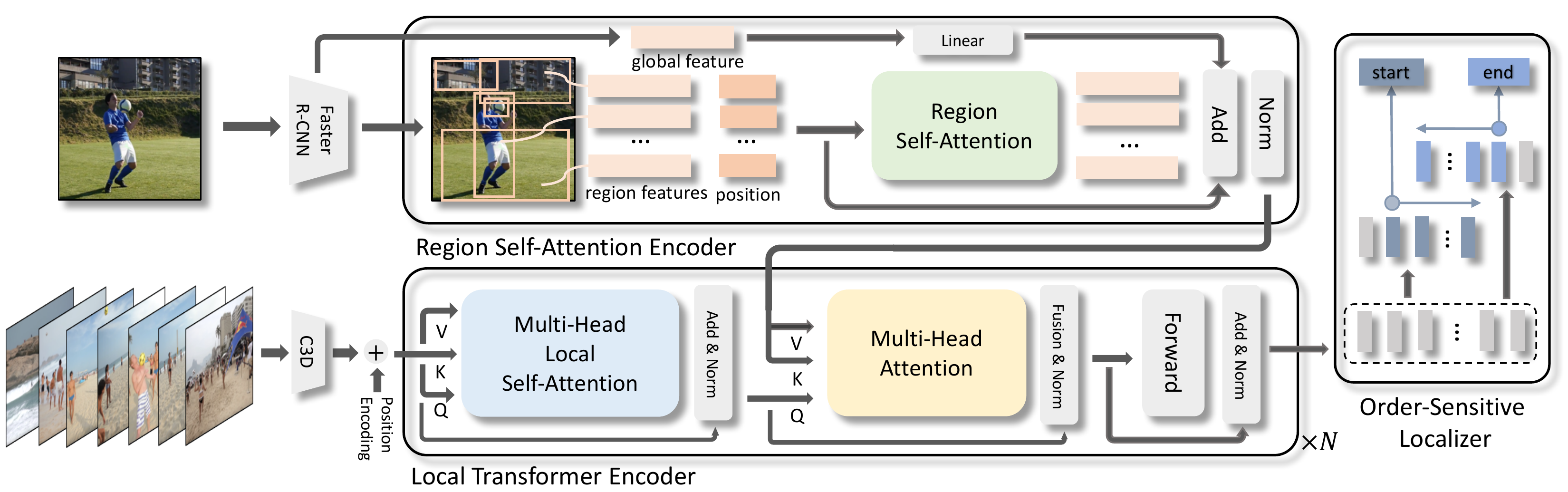}
\caption{The Framework of Self-Attention Interaction Localizer for Image-Based Activity Localization.}\label{fig:framework}
\end{figure*}

\section{Preliminary}
In this section, we introduce the scaled dot-product attention and multi-head attention~\cite{vaswani2017attention}.
% , which are proposed in the original Transformer architecture

We assume the input of the scale dot-product attention is a sequence of queries ${\bf Q} \in \mathbb{R}^{d_1 \times n_1}$, keys ${\bf K} \in \mathbb{R}^{d_1 \times n_2} $ and values ${\bf V} \in \mathbb{R}^{d_2 \times n_2}$, where $n_1$ , $n_2$ and $n_2$ represent the number of queries, keys and values, and $d_1$, $d_1$  and $d_2$ are respective dimensions,
the scaled dot-product attention is then calculated by
\begin{eqnarray}
& {\rm DotAtten}({\bf Q},{\bf K},{\bf V}) = {\rm Softmax}({\bf Q}^{\top}{\bf K}}/{\sqrt{d_1}){\bf V}^{\top}, \nonumber
\end{eqnarray}
where the $ {\rm Softmax}$ operation is performed for every row.

The multi-head attention consists of $H$ paralleled units. Each unit first projects the queries, keys and values to $d_1$, $d_1$ and $d_2$ dimenstions, then independently perform the scale dot-product attention, and finally concatenate their outputs to another linear projection, given by
\begin{eqnarray}
&{\rm MultiHead}({\bf Q},{\bf K},{\bf V}) = {\bf W}^o{\rm Concat(head_1,\ldots, head_H)} \nonumber \\
&{\rm where \ head_i = DotAtten({\bf W}_i^q{\bf Q}, {\bf W}_i^k{\bf K}, {\bf W}_i^v{\bf V})}, \nonumber
\end{eqnarray}
where ${\bf W}_i^q \in \mathbb{R}^{d_1 \times d_q}$, ${\bf W}_i^k \in \mathbb{R}^{d_1 \times d_k}$, ${\bf W}_i^v \in \mathbb{R}^{d_2 \times d_v}$, ${\bf W}^o \in \mathbb{R}^{d_v \times Hd_2}$ are linear projection matrices. The $d_q$, $d_k$ and $d_v$ are the initial input dimensions and $d_v$ is the output dimension. If the queries, keys and values are set to the same sequences, it is called multi-head self-attention.

\section{Self-Attention Interaction Localizer}

\subsection{Problem Formulation}
We denote a video by ${\bf v} = \{{\bf v}_{i}\}_{i=1}^{n}$, where ${\bf v}_i$ is the feature of the $i$-th frame and $n$ is the frame number. Each video is associated with an image query ${\bf q} $, which corresponds to a video segment depicting the similar semantic contents. And the start and end boundaries of this target segment are denoted by  ${\bf \tau} = (s, e)$.
Thus, given a set of videos, image queries and segment annotations, the goal of the self-attention interaction localizer is to predict the boundaries ${\hat \tau} = ({\hat s}, {\hat e})$ of unseen activities during the inference process. The overall framework is shown in Figure~\ref{fig:framework}.

\subsection{Region Self-Attention Encoder}
In this section, we introduce the region self-attention encoder with relative position encoding to learn fine-grained image region representations.

The image query often contains a series of objects in different regions, and the interactions between these objects imply crucial semantic information. Thus, we develop the region self-attention to explore the region relations for fine-grained understanding.
Similar to the region proposal~\cite{Anderson2018BottomUpAT} for image caption, we first extract basic region features from the image query using a pre-trained Faster R-CNN~\cite{ren2015faster}, denoted by ${\bf q}=({\bf r}_{1},{\bf r}_{2},\ldots,{\bf r}_{m})$ of $m$ regions. Each region ${\bf r}_i$ contains an object and associates with a 4-dimensional position vector ${\bf p}_i = (x_{i},y_{i}, w_{i}, h_{i})$, where $x_i$ and $y_i$ are center coordinates, and $w_i$ and $h_i$ represent width and height of this region. Moreover, we pool the overall feature maps of the last shared convolutional layer of Faster R-CNN to obtain the global image feature ${\bf f}^g$. 

\begin{figure}[t]
\centering
\includegraphics[width=0.48\textwidth]{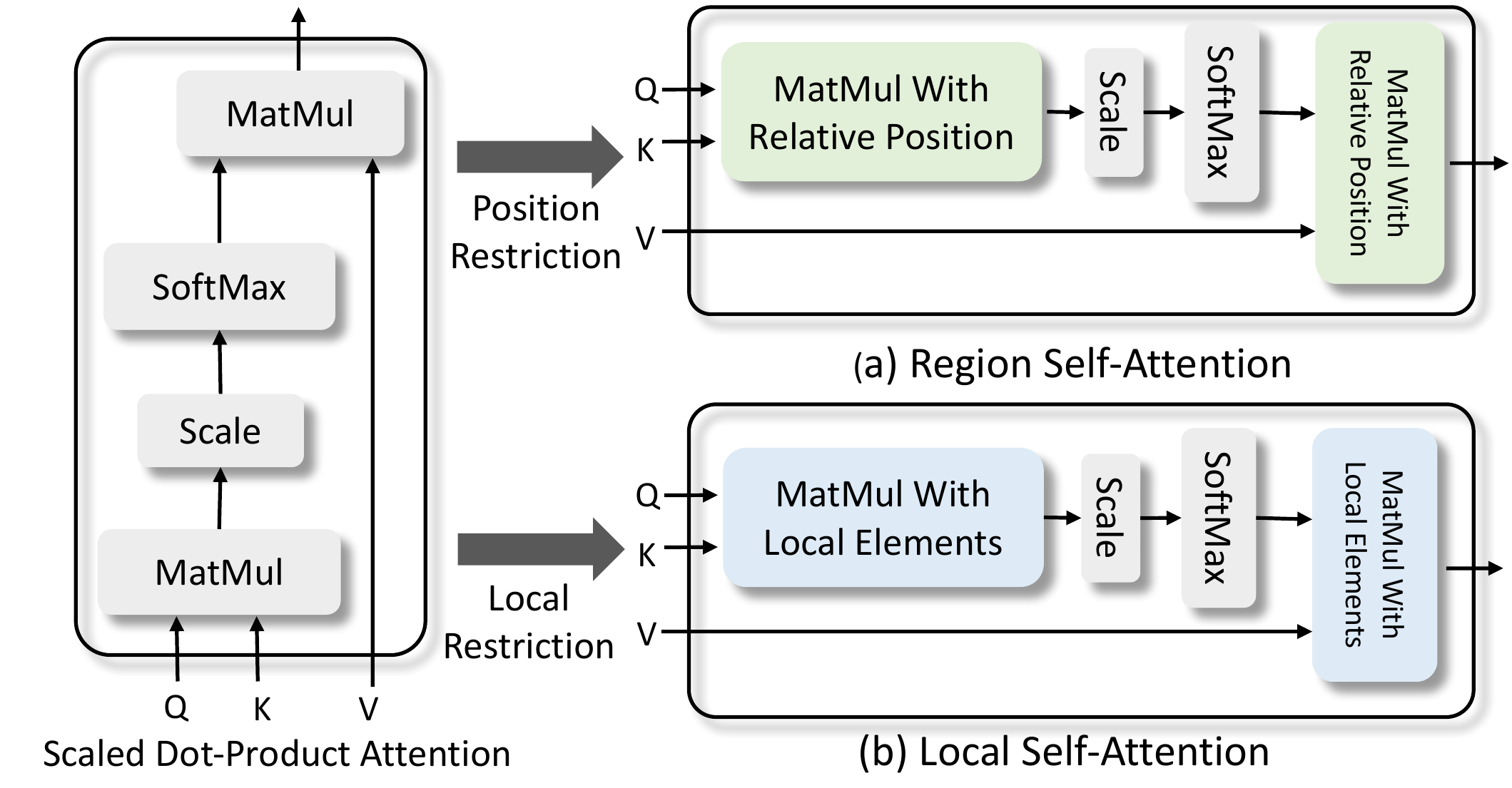}
\caption{The variants of scaled dot-product attention. (a) The region self-attention contains relative position encoding. (b) The local self-attention only focuses on its surrounding elements.}\label{fig:framework2}
\end{figure}

We then develop the region self-attention method to establish the potential relations between objects. 
The original self-attention approach~\cite{vaswani2017attention} is based on scaled dot-product attention. 
Setting ${\bf Q}$, ${\bf K}$ and ${\bf V}$ to the same region matrix ${\bf R} = [{\bf r}_1;{\bf r}_{2};\ldots;{\bf r}_{m}] \in \mathbb{R}^{d_r \times m}$, the ${\rm DotAtten({\bf Q},{\bf K},{\bf V})}$ becomes a self-attention operation, where $d_r$ is the dimension of region features. Furthermore, we consider the relative position of each region pair to build more informative interactions, shown in Figure~\ref{fig:framework2}(a). Specifically, given the position vector ${\bf p}_i$ and ${\bf p}_j$ of the $i$-th and $j$-th regions, the relative position vector ${\bf p}_{ij}^{r} = (x_{ij},y_{ij}, w_{ij}, h_{ij})$  is computed by
\begin{eqnarray}
&x_{ij} = (x_i - x_j)/w_j, \ y_{ij} = (y_i - y_j)/h_j,  \nonumber\\
&w_{ij} = {\rm log}(w_i/w_j), \ h_{ij} = {\rm log}(h_i/h_j). \nonumber
\end{eqnarray}
We then project the 4-dimensional vector to $d_r$ dimension, denoted by ${\bf \widetilde p}^r_{ij} = {\bf W}^r {\bf p}_{ij}^{r}$, where ${\bf W}^r \in \mathbb{R}^{d_r \times 4}$ represents the projection weight.
Thus, for each region ${\bf r}_i$, we can obtain a relative position matrix $ {\bf \widetilde P}_i = [{\bf \widetilde p}^r_{i1}; {\bf \widetilde p}^r_{i2};\ldots;{\bf \widetilde p}^r_{im}] \in \mathbb{R}^{d_r \times m}$, and the region self-attention result for the $i$-th region is given by
\begin{eqnarray}
{\bf \widetilde h}_i^r = {\rm Softmax}({\bf r}_i^{\top} ({\bf R}+{\bf \widetilde P}_i)}/{\sqrt{d_r})({\bf R}+{\bf \widetilde P}_i)^{\top}, \nonumber
\end{eqnarray}
where the region representation ${\bf \widetilde h}_i^r$ incorporates relevant information from other regions with relative position encoding. Next, we employ a residual connection~\cite{he2016deep} to keep basic region information, and add the global feature to each region by a linear projection, followed by a layer normalization~\cite{ba2016layer}:
\begin{eqnarray}
{\bf  h}_i^r = {\rm LayerNorm}({\bf \widetilde h}_i^r + {\bf r}_i +  {\bf W}_g{\bf f}^g). \nonumber
\end{eqnarray}
Hence, by the region self-attention encoder, we finally obtain the fine-grained region representations ${\bf H}^{r} \in \mathbb{R}^{d_r \times m}$, consisting of $[{\bf h}^{r}_{1};{\bf h}^{r}_{2},\ldots;{\bf h}^{r}_{m}]$.

\subsection{Local Transformer Encoder}
In this section, we propose a local transformer encoder to build multi-step fusion and reasoning of image and video contents, where multi-level interactions can sufficiently explore the query semantics for fuzzy localization.

Following the original transformer framework~\cite{vaswani2017attention}, we first add a temporal information encoding to the frame sequences $\{{\bf v}_{i}\}_{i=1}^{n}$, then develop the $K$-layer local transformer encoder, where the output of $k-1$-th layer is regarded as the input of the $k$-th layer. Each layer consists of a multi-head local self-attention sublayer, a multi-head attention sublayer and a feed-forward sublayer. And the residual connection is applied from input to output of each sublayer, followed by layer normalization. Next, we detailedly illustrate the $1$-th layer of our encoder, which takes the initial frame sequence $\{{\bf v}_{i}\}_{i=1}^{n}$ as input.

We first introduce the multi-head local self-attention sublayer. Given the feature matrix ${\bf F} = [{\bf v}_1;{\bf v}_{2};\ldots;{\bf v}_{n}] \in \mathbb{R}^{d_f \times n}$ with $n$ elements of $d_f$ dimension, the original multi-head self-attention ${\rm MultiHead}({\bf F},{\bf F},{\bf F})$ captures global frame-to-frame dependencies. However, retrieving the target segment for image queries mainly depends on local relation modeling. Thus, we augment the local restrictions for the underlying scaled dot-product attention, shown as Figure~\ref{fig:framework2}(b). For the $i$-th frame, the local self-attention only focuses on its surrounding elements, given by
\begin{eqnarray}
{\bf \widetilde v}_i = {\rm Softmax}({\bf v}^{\top}_i{\bf F}_{:,i-w:i+w}}/{\sqrt{d_f}){\bf F}_{:,i-w:i+w}^T, \nonumber
\end{eqnarray}
where the $w$ is the window size of local self-attention and ${\bf F}_{:,i-w:i+w}$ represents consecutive $2w+1$ frames around the $i$-th element. This local strategy is similar to the convolutional neural networks, but CNNs share the convolution kernel at each time step, while our local self-attention relies on the current element. That is, the CNNs are location-based modeling and our local self-attention belongs to content-based modeling.
Furthermore, we apply the multi-head setting for local self-attention just like the multi-head dot-product attention, called ${\rm LocalMultiHead}$, and obtain the output of the $1$-th sublayer by ${\bf F}^{11} = {\rm LayerNorm} ({\rm LocalMultiHead}({\bf F},{\bf F},{\bf F})+{\bf F})$.

We then propose a multi-head attention sublayer over the image region representations ${\bf H}^{r}$. 
Specifically, given ${\bf H}^{r}  = [{\bf h}^{r}_{1},{\bf h}^{r}_{2},\ldots,{\bf h}^{r}_{m}] \in \mathbb{R}^{d_r \times m} $ and ${\bf F}^{11} \in \mathbb{R}^{d_f \times n}$, we aggregate the frame-aware image representations by 
${\bf \widetilde F}^{11} = {\rm MultiHead}({\bf F}^{11},{\bf H}^{r}, {\bf H}^{r})$,
where ${\bf \widetilde F}^{11}$ incorporates the crucial query information for each frame and eliminates the influence of semantically inessential query contents.
After that, we apply a more informative fusion~\cite{mou2015natural} instead of the additive residual connection, given by 
\begin{eqnarray}
& {\rm Fusion}({\bf F}^{11}_{:,i},{\bf \widetilde F}^{11}_{:,i} ) = {\rm tanh}({\bf W}_f{\bf \overline F}^{11}_{:,i} + {\bf b}_f), \nonumber \\
& {\rm where} \ \ {\bf \overline F}^{11}_{:,i} = [{\bf F}^{11}_{:,i}; {\bf \widetilde F}^{11}_{:,i}; {\bf F}^{11}_{:,i} \odot {\bf \widetilde F}^{11}_{:,i};{\bf F}^{11}_{:,i} - {\bf \widetilde F}^{11}_{:,i}], \nonumber
\end{eqnarray}
where $\odot$ represents the element-wise product and $[\cdot ; \cdot]$ is the feature concatation. Thus, we get the output of the $2$-th sublayer by ${\bf F}^{12} = {\rm LayerNorm} ({\rm Fusion}( {\bf F}^{11} ,{\bf \widetilde F}^{11}))$.

In addition to the two attention sublayers, each layer contains a feed-forward sublayer, composed of two linear transformations with ReLU activation~\cite{nair2010rectified}:
\begin{eqnarray}
& {\bf \widetilde F}^{12} = {\bf W}^2_{13} \ {\rm ReLU}({\bf W}^1_{13} {\bf F}^{12}+{\bf b}^1_{13} {\bf 1}^{\top})+{\bf b}^2_{13} {\bf 1}^{\top}. \nonumber
\end{eqnarray}
At the end, we obtain the output of the $1$-th layer by ${\bf F}^{13} = {\rm LayerNorm} ({\bf F}^{12} + {\bf \widetilde F}^{12})$.

By stacking $N$ layers, our local transformer encoder builds multi-step fusion and reasoning of image and video contents, and obtain the final semantic representations ${\bf F}^{N3}$, also denoted by ${\bf H}^{v} = [{\bf h}^{v}_{1};{\bf h}^{v}_{2};\ldots;{\bf h}^{v}_{n}]$.

\subsection{Order-Sensitive Localizer}
In this section, we apply an order-sensitive localizer to directly retrieve the start and end boundaries of the target segment, which leverages the bi-directional contextual information to boost the localization accuracy.

While predicting the start and end boundaries, the information from different directions has different influences~\cite{chen2019localizing}. For example, we suppose the $i$-th frame is the start frame of the target activity, the forward information after this frame can reflect whether a complete activity just starts at the $i$-th frame, and vice versa. 
Based on this observation, we propose a bi-directional aggregation method. Specifically, we first aggregate forward semantic representations for the $i$-th frame, given by 
\begin{eqnarray}
&{\bf h}^{fw}_i = \sum^n_{t = i} \alpha^{fw}_t {\bf h}^v_t, \  \alpha^{fw}_{it} = \frac{exp({a}^{fw}_{it})}{\sum^n_{t = i} exp({a}^{fw}_{it})}, \nonumber \\
&{a}^{fw}_{it} = {\bf w}^T_{fw} {\rm tanh}({\bf W}_{fw}^1 {\bf h}^v_i + {\bf W}_{fw}^2 {\bf h}^v_t),  \nonumber 
\end{eqnarray}
where ${\bf h}^{fw}_i$ is the forward context representation of the $i$-th frame and $\alpha^{fw}_{it}$ is the attention weight from the $i$-th frame to the $t$-th frame, where $t \geq i$. Instead of preceding dot-product attention, we apply the additive attention here, which is slower but more effective. 
Likewise, the backward context representation ${\bf h}^{bw}_i$ is generated by aggregating those frame representations before the index $i$.
Given the bi-directional context representation ${\bf H}^{fw} = [{\bf h}^{fw}_1;{\bf h}^{fw}_2;\ldots;{\bf h}^{fw}_n]$  and ${\bf H}^{bw} = [{\bf h}^{bw}_1;{\bf h}^{bw}_2;\ldots;{\bf h}^{bw}_n]$ , we can directly predict the start and end boundaries as follows
\begin{eqnarray}
&{\bf p}^{s} = {\rm Softmax}({\bf w}^T_{s}{\rm tanh}({\bf W}_s {\bf H}^{fw} + {\bf b}_s {\bf 1}^{\top})), \nonumber \\
&{\bf p}^{e} = {\rm Softmax}({\bf w}^T_{e}{\rm tanh}({\bf W}_e {\bf H}^{bw} + {\bf b}_e {\bf 1}^{\top})),  \nonumber \\
& {\hat s} = \underset{i}{\rm arg \ max}({\bf p}^{s}_i), \ {\hat e} = \underset{i}{\rm arg \ max}({\bf p}^{e}_i). \nonumber
\end{eqnarray}
where ${\bf p}^{s}$ and ${\bf p}^{e}$ are the probability distributions of start and end boundaries over frame sequences, and the ${\hat \tau} = ({\hat s}, {\hat e})$ is the retrieval result. Finally, we train our self-attention interaction localizer by minimizing the negative log probabilities of the ground truth boundaries by
\begin{eqnarray}
&{\mathcal L} = -\frac{1}{N_b}\sum\limits_{1}^{N_b}({\rm log}{\bf p}^s_{ s}+{\rm log}{\bf p}^e_{e}), \nonumber
\end{eqnarray}
where ${\bf p}^s_{ s}$ is the probability of ground truth start, ${\bf p}^e_{e}$ is the end probability and $N_b$ is the sample numbers of a batch.

\section{Dataset}
Since none of existing datasets can be directly used for image-based activity localization, we construct a new dataset ActivityIBAL by reorganizing the ActivityNet~\cite{caba2015activitynet}.
The AcitivityNet is a large-scale video dataset for action localization, which provides 14950 public videos with temporal annotations of 200 classes. Videos in ActivityNet contain one or multiple action segments, and these segments in one video often have the same action label. 

Following similar reconstruction~\cite{feng2018video}, we preprocess these videos as follows. We first merge every two overlapped segments belonging to the same labels in videos, and then split each original video with multiple action segments into several independent videos. Next, we discard some videos with too short or long action segments, and divide these newly generated videos into three subsets according to their action classes: 160 classes for training, 20 classes for validation and the remaining 20 classes for testing. This setting guarantees these videos used in validation and testing do not appear during the training process. 

We then assume that the segments belonging to the same class describe the semantically related activities. 
The ActivityIBAL dataset contains two parts of data: we first construct the {\bf Simple} part, where an untrimmed video corresponds to an image query selected from another video with the same class; then select the image query from the similar but not same class to compose the {\bf Difficult} part. Here we manually select the most informative frame as the image query.  The similar class can be found in the tree structure of these classes in ActivityNet.
Note that, to keep the diversity of image queries, we construct 5 samples for every video, corresponding to 3 simple queries and 2 difficult queries. 
The details are summarized in Table~\ref{table:activitydataset}.

\begin{table}[t]
\centering
\begin{tabular}{c|cccc}
\hline
     &Sample& Class &  Video Time&   Target Time\\
\hline
\hline
Train&30,677&160 &63.17&22.37\\
\hline
Valid&2,995&20 &73.38&25.21\\
\hline
Test&3,013&20 &61.20&21.61\\
\hline
All&36,685 & 200&63.84&22.54\\
\hline
\end{tabular}
\caption{Summaries of the ActivityIBAL Dataset, including sample number , class number, average video duration and average segment duration.}\label{table:activitydataset}
\end{table}

\section{Experiments}
\subsection{Implementation Details}
For image queries, we first extract the 4,096-d region feature and 4-d position vector for each region using a pre-trained Faster R-CNN~\cite{ren2015faster} with VGG-16 backbone networks, and then apply the non-maximum suppression with threshold 0.7 to reduce redundant regions. 
By the way, we obtain the 512-d global feature.
As for video sequences, we first resize each frame to 112$\times$112, and then extract 4,096-d spatio-temporal features by a pre-trained 3D-ConvNet~\cite{tran2015learning} for each unit. A unit contains consecutive 16 frames and overlaps 8 frames with adjacent units. Next, we reduce the dimension from 4,096 to 500 using PCA. To avoid exceeding GPU load, we downsample overlong features sequences to 200. 
In the proposed model, we set the projection dimension of various kinds of attention to 256.
During the training process, we adopt an Adam optimizer and the initial learning rate is set to 0.0005.

\subsection{Baselines and Evaluation Criteria}
Since these are no other approaches designed for image-based activity localization, we compare our SAIL method with three baselines, one designed by ourselves and two extended from relevant tasks.
\begin{itemize}
\item \textbf{FLP}: We design a simple baseline to perform the frame-level localization. For each frame, given frame feature ${\bf v}_i$ and global image feature ${\bf f}^g$, we compute the confidence score by multilayer perceptron $ p_i = {\rm MLP}({\bf v}_i, {\bf f}^g)$. The score of each segment $(i,j)$ is given by $p_{ij} = \prod_{k=1}^{i-1}(1-p_k)\prod_{k=i}^{j}(p_k)\prod_{k=j+1}^{n}(1-p_k)$.
\item \textbf{SST+}~\cite{buch2017sst}: The baseline is the extension of an action proposal method (SST). We add a fusion layer on the top of original GRU networks to incorporate the global image feature, and then choose the proposal with the largeset confident score.
\item \textbf{TAL+}~\cite{chao2018rethinking}: Based on an action localization method, we fisrt generate some action proposals, then add an attention layer to aggregate the image region features for each proposal and finally estimate the proposal score.
\end{itemize}

\begin{table}[t]
\centering
\begin{tabular}{c|ccccccc}
\hline
Method &mIoU&   IoU@0.3&    IoU@0.5  & IoU@0.7\\
\hline
\hline
Random&14.68\%&18.13\%&10.28\%&6.87\%\\
FLP&23.73\%&29.81\%&19.49\%&12.37\%\\
SST+&40.63\%&59.92\%&32.17\%&14.50\%\\
TAL+&41.93\%&64.55\%&36.50\%&17.86\%\\
\hline
SAIL&{\bf 49.66\%}&{\bf73.98\%}&{\bf48.25\%}&{\bf24.37\%} \\
\hline
\end{tabular}
\caption{Evaluation Results on the whole ActivityIBAL Dataset.}\label{table:result}
\end{table}

\begin{table}[t]
\centering
\begin{tabular}{c|ccccccc}
\hline
Method &mIoU&   IoU@0.3&    IoU@0.5  & IoU@0.7\\
\hline
\hline
Random&14.32\%&18.22\%&10.19\%&6.77\%\\
FLP&17.89\%&22.30\%&14.85\%&9.42\%\\
SST+&32.37\%&48.06\%&26.79\%&11.26\%\\
TAL+&35.20\%&56.44\%&32.05\%&13.39\%\\
\hline
SAIL&{\bf 45.71\%}&{\bf68.23\%}&{\bf45.17\%}&{\bf21.84\%} \\
\hline
\end{tabular}
\caption{Evaluation Results on the Difficult Part of AcitivityIBAL.}\label{table:result2}
\end{table}

\textbf{Evaluation Criteria}: We evaluate the performance of all methods on the criteria {\bf mIoU} and {\bf IoU@R}.
Specifically, we first calculate the Intersection over Union (IoU) for each pair of predicted segments and ground-truth ones, and then define the {\bf mIoU} is a mean IoU of all testing samples. And {\bf IoU@R} represents the percentage of the samples, which IoU $>$ R.

\begin{figure}[t]
\centering
\scalebox{0.99}{
\subfigure[mIoU]{\includegraphics[width=0.24\textwidth]{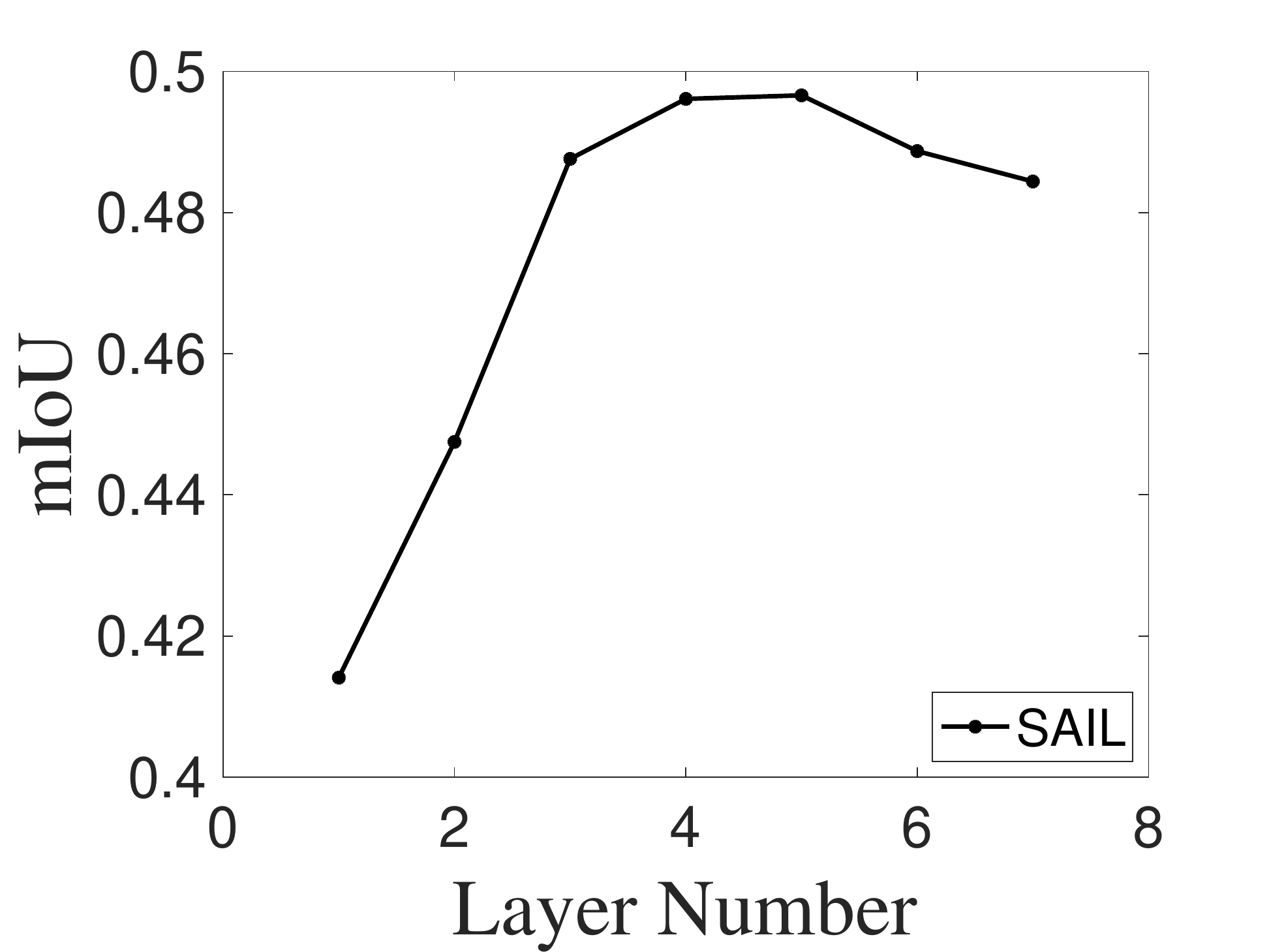}}
\subfigure[IoU@0.3]{\includegraphics[width=0.24\textwidth]{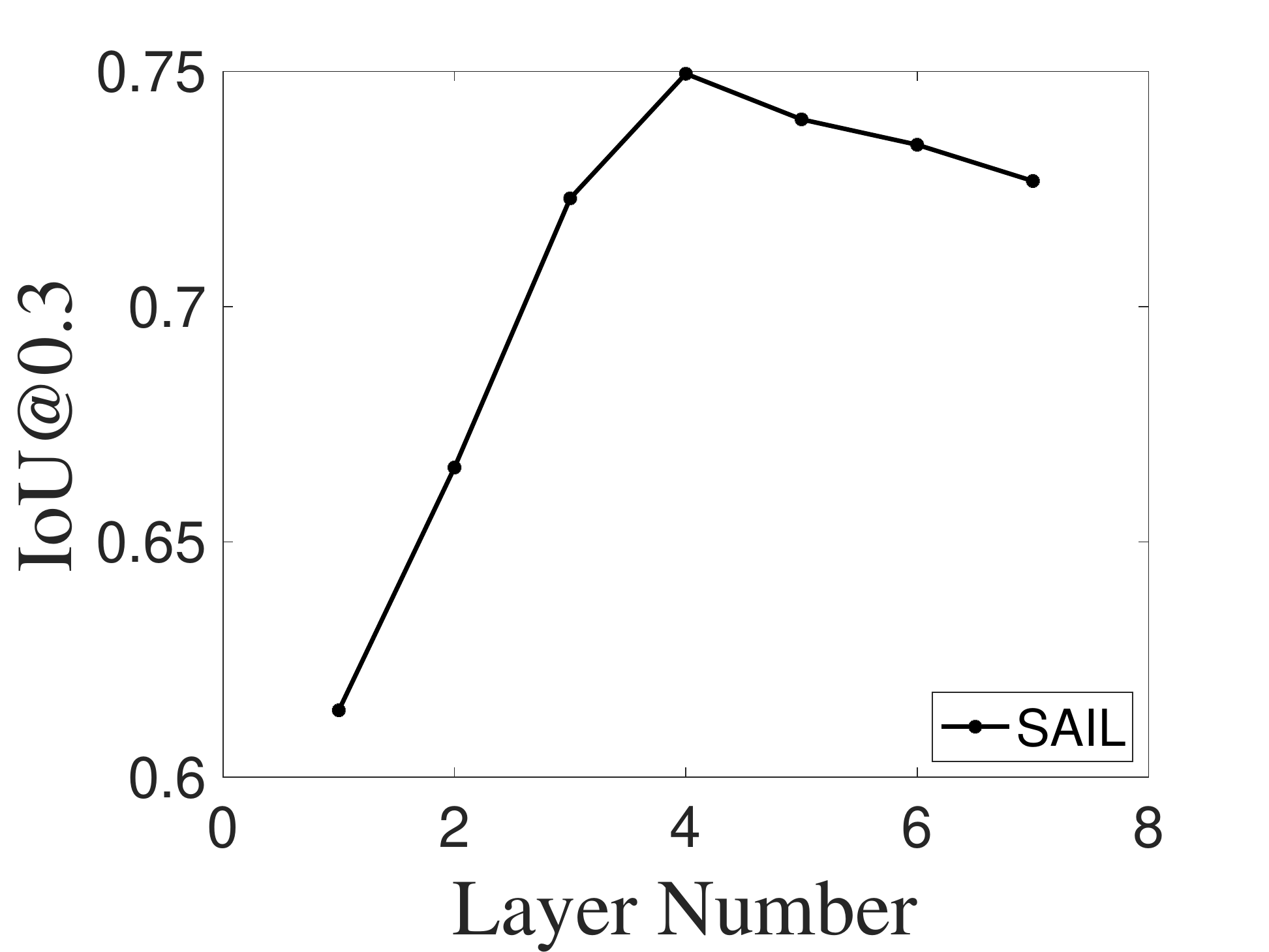}}
}
\scalebox{0.99}{
\subfigure[IoU@0.5]{\includegraphics[width=0.24\textwidth]{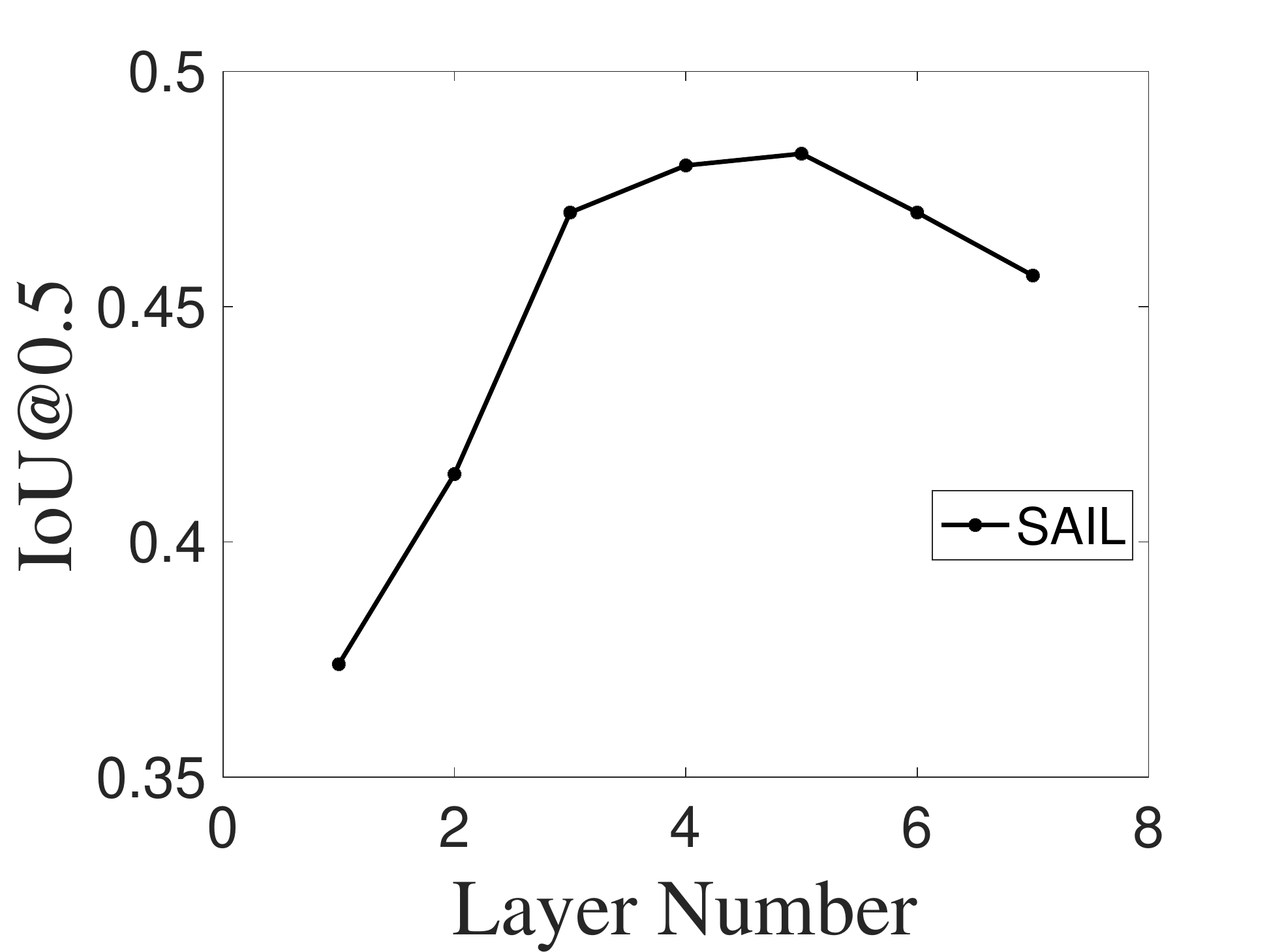}}
\subfigure[IoU@0.7]{\includegraphics[width=0.24\textwidth]{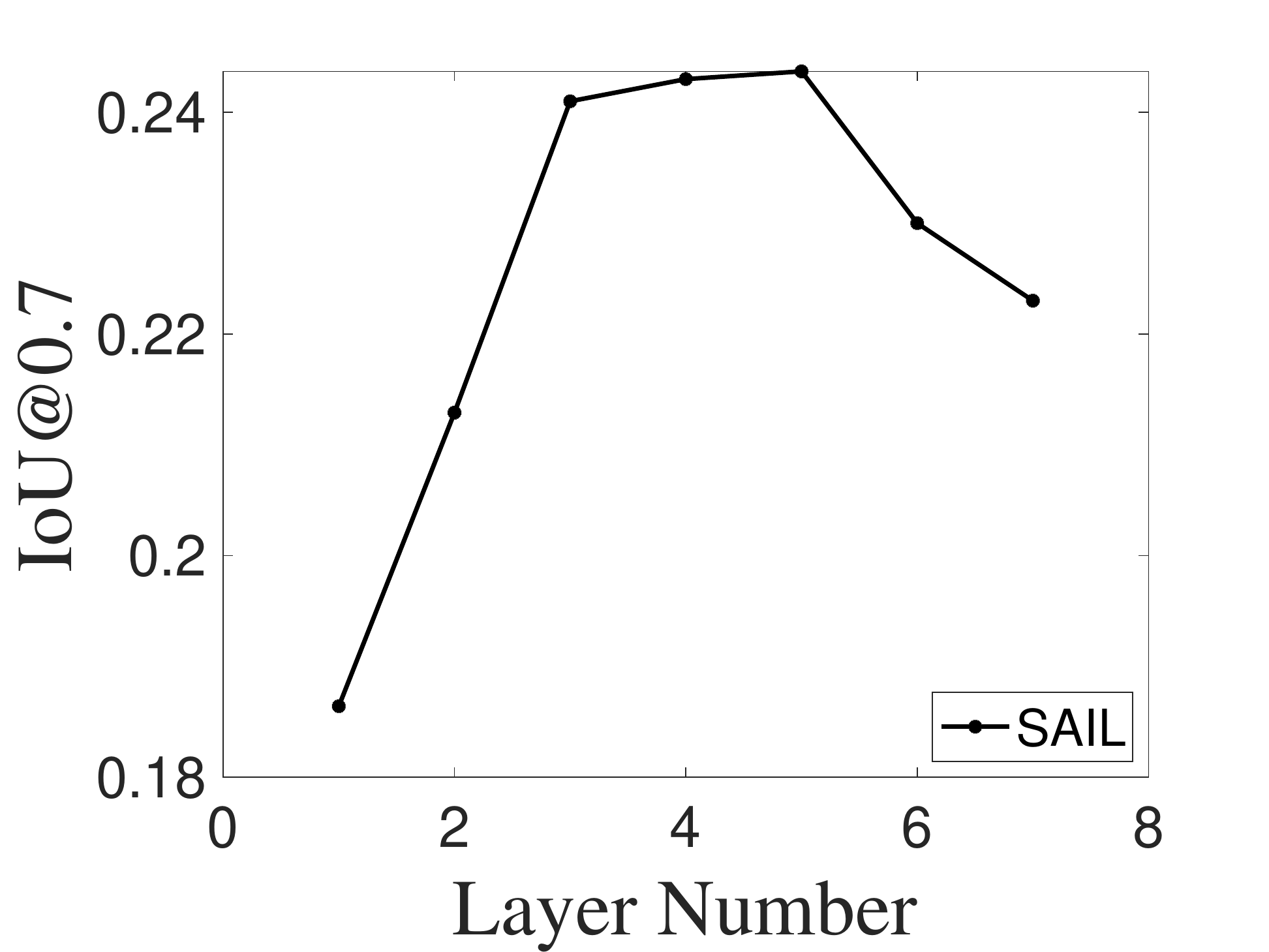}}
}
\caption{Effect of layer number of the local transformer encoder.}\label{fig:layer}
\end{figure}

\subsection{Performance Evaluation}
\begin{figure}[t]
\centering
\includegraphics[width=0.48\textwidth]{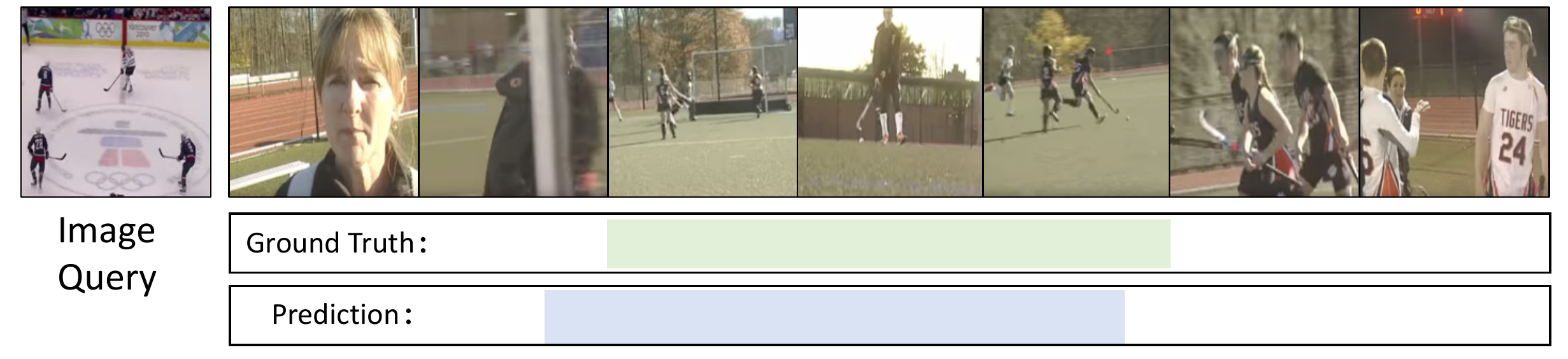}
\caption{A typical example of image-based activity localization.}\label{fig:attention}
\end{figure}

We then evaluate all models on the ActivityIBAL dataset. 
The overall results are shown in Table~\ref{table:result}, where the random results are yielded by selecting a segment randomly. And the Table~\ref{table:result2} shows the performance only on the difficult testing samples of ActivityIBAL, where the training process is still performed on the whole training set.
The experiment results reveal some interesting points:
\begin{itemize}
\item The frame-level FLP method achieves the worst performance on all criteria, and is only a little better than the random strategy. It demonstrates that the image-based activity localization cannot only depend on the frame-level modeling.
\item The attention based method SAIL and TAL+ outperform other baselines, which suggests that the frame-aware aggregation of image region features is critical to localize unseen activities.
\item In all the criteria, our SAIL method achieves the best performance. This fact shows that our proposed region self-attention encoder, local transformer encoder and order-sensitive localizer can effectively improve the localization accuracy of unseen activities.
\item Although all methods have lower accuracy on the Difficult samples than the whole testing set, our SAIL still outperforms other baselines and has less performance degradation, which demonstrates our strategy has the stronger ability of fuzzy localization.
\end{itemize}
In our SAIL method, the stacked layer number of the local transformer encoder is an essential hyper-parameter. We next explore the impact of this hyper-parameter. Concretely, we vary the layer number from 1 to 7 and display the results in Figure~\ref{fig:layer}. From these figures, we note that too many or too few layers will both affect the performance, and our SAIL method achieves the best performance when the layer number is set to 5.
Moreover, we display a typical example of image-based activity localization in Figure~\ref{fig:attention}, where we localize an unseen activity ``grass hockey'' by an image ``ice hockey''.{}
\begin{table}[t]
\centering
\begin{tabular}{c|cccccc}
\hline
Method  & mIoU & IoU@0.3&   IoU@0.5 &  IoU@0.7 \\
\hline
\hline
w/o. RS &46.63\%&70.35\%&43.36\%&20.43\%\\
w/o. ML &48.40\%&73.29\%&44.07\%&21.24\%\\
w/o. LS &47.35\%&71.17\%&45.71\%&22.29\%\\
w/o. BA &48.04\%&72.00\%&45.76\%&23.60\%\\
\hline
full&{\bf 49.66\%}&{\bf 73.98\%}&{\bf48.25\%}&{\bf24.37\%}\\
\hline
\end{tabular}
\caption{Evaluation Results of Ablation Study.}\label{table:ablation}
\end{table}

\subsection{Ablation Study}
We next verify the contribution of each component by the ablation study.
We first remove the region self-attention from the image encoder, denoted by {\bf SAIL(w/o. RS)}. 
To validate the effect of multi-level interactions, {\bf SAIL(w/o. ML)} discards the multi-head attention sublayer from the first $N-1$ layers of the local encoder. We next replace the multi-head local self-attention with the normal multi-head self-attention, called as {\bf SAIL(w/o. LS)}. Finally, we develop a model {\bf SAIL(w/o. BA)} without the bi-directional aggregation method in the order-sensitive localizer.

As shown in Table~\ref{table:ablation}, the SAIL(full) outperforms all ablation models, demonstrating the region self-attention, multi-level interactions, local self-attention and bi-directional aggregation method are all effective to boost the localization performance. 
Specifically, SAIL(w/o. RS) achieves the worst performance, indicating the fine-grained region representation learning is the basis of accurate localization.
Moreover, SAIL(w/o. ML) has a more severe performance degradation on IoU@0.5 and IoU@0.7 than mIoU and IoU@0.3, demonstrating the multi-level interactions are more effective for accurate localization.
On the contrary, we note that the local self-attention and bi-directional aggregation method has a greater impact on the rough localization.

\section{Conclusion}
In this paper, we introduce a new task to localize unseen activities in videos via image queries. To deal with the inherent challenges of this task, we propose a novel self-attention interaction localizer.Furthermore, we reorganize the ActivityNet to construct a new dataset ActivityIBAL. 
The extensive experiments show the effectiveness of our method.

\section*{Acknowledgments}
This work was supported by the National Natural Science Foundation of China under Grant No.61602405, No.U1611461, No.61751209 and No.61836002, Joint Research Program of ZJU and Hikvision Research Institute.
This work is also supported by Alibaba Innovative Research and Microsoft Research Asia.

%% The file named.bst {}is a bib{}liography style file for BibTeX 0.99c
\clearpage
\bibliographystyle{named}
\bibliography{ijcai19}

\end{document}